\documentclass[lettersize,journal]{IEEEtran}
\usepackage{amsmath,amsfonts}
\usepackage{algorithm}
\usepackage{array}
\usepackage[caption=false,font=normalsize,labelfont=sf,textfont=sf]{subfig}
\usepackage{textcomp}
\usepackage{stfloats}
\usepackage{url}
\usepackage{verbatim}
\usepackage{graphicx}
\usepackage{cite}
\usepackage{multirow}
\usepackage{xcolor}

\usepackage{longtable}
\usepackage{booktabs}
\usepackage{array}
\usepackage{tabularx} 
\usepackage{utfsym}
\usepackage{colortbl}
\usepackage{xcolor}
\usepackage{array}
\usepackage{color}

\usepackage{pifont}

\usepackage{soul}
\usepackage{amssymb}
\usepackage{bbding}
\usepackage{subcaption}

\usepackage{colortbl}
\usepackage{newfloat}
\usepackage{listings}

\usepackage{epsfig}
\usepackage{caption}
\usepackage{setspace}
\usepackage{algpseudocode}
\usepackage{tabu}

\begin{document}

\title{Referring Video Object Segmentation with Cross-Modality Proxy Queries}

\author{Baoli Sun, Xinzhu Ma, Ning Wang, Zhihui Wang, Zhiyong Wang
        % <-this % stops a space
\thanks{This work was supported by Dalian High-Level Talents Innovation Support Plan under Grant 2021RQ052, and Australian Research Council (ARC) Discovery Project under Grant DP210102674. (Corresponding author: Zhihui Wang, Email: zhwang@dlut.edu.cn.)}% <-this % stops a space
\thanks{B. Sun, N. Wang, Z. Wang  are with DUT-RU International School of Information Science \& Engineering, Dalian University of Technology, China.}
\thanks{X. M is with the Chinese University of Hong Kong, China.}
\thanks{Z. Wang is with the University of Sydney, Australia.}
}

% The paper headers
\markboth{IEEE TRANSACTIONS ON MULTIMEDIA}
{Baoli Sun \MakeLowercase{\textit{et al.}}: Referring Video Object Segmentation with Cross-Modality Proxy Queries}

\maketitle

\begin{abstract}
Referring video object segmentation (RVOS) is an emerging cross-modality task that aims to generate pixel-level maps of the target objects referred by given textual expressions. The main concept involves learning an accurate alignment visual elements and language expressions within a semantic space. Recent approaches address cross-modality alignment through conditional queries, tracking  the target object using a query-response based mechanism built upon transformer structure. However, they exhibit two limitations: (1) these conditional queries, identifying the same object across different frames through the same query, lack inter-frame dependency and variation modeling, making accurate target tracking challenging amid significant frame-to-frame variations;
and (2) they handle the temporal feature of a video and build visual-language interaction sequentially, integrating textual constraints belatedly,  which may cause the video features potentially focus on the non-referred objects. 
Therefore, we propose a novel RVOS architecture called ProxyFormer, which introduces a set of proxy queries to integrate visual and text semantics and facilitate the flow of semantics between them.
By progressively updating and propagating proxy queries across multiple stages of video feature encoder, ProxyFormer ensures that the video features are as focused as much possible on the object of interest. 
This dynamic evolution of the queries across video  also enables the proxy queries to establish inter-frame dependencies, enhancing the accuracy and coherence of object tracking throughout the video sequence.
To mitigate the high computational costs associated with full spatio-temporal interactions between video  and proxy queries, we propose to decouple cross-modality interactions into their temporal and spatial dimensions, respectively. 
Additionally, we design a Joint Semantic Consistency (JSC) training strategy to align semantic consensus between the proxy queries and the combined video-text pairs.
Comprehensive experiments on four widely used RVOS benchmarks, \textit{i.e}., \textit{Ref-Youtube-VOS, Ref-DAVIS17, A2D-Sentences} and \textit{JHMDB-Sentences}, clearly demonstrate the superiority of our ProxyFormer to the state-of-the-art methods.
\end{abstract}


\section{Introduction}
\label{sec:intro}
Referring video object segmentation (RVOS) is an emerging field that involves segmenting and tracking target objects in videos based on the given natural language expressions. This area has attracted significant interest from the research community due to its potential to enhance various applications, such as video editing, virtual reality, and human-robotic interaction. Different from the traditional single-video VOS tasks, which rely on pre-defined categories  or visual cues, RVOS demands a thorough comprehension of the content across different modalities to accurately identify and segment target objects. Thus, the inherent challenge of RVOS  lies in establishing the visual-language alignment, bridging the compact semantics of language expressions with the long spatio-temporal semantics in video content.
%between the compact semantic of language expression and long-term spatio-temporal semantic of video.
%the ambiguous temporal  semantics in  language expressions and 
%%%

\begin{figure*}[!h]
	\centering
        %\vspace{-6pt}
	\centerline{\includegraphics[width=1\linewidth]{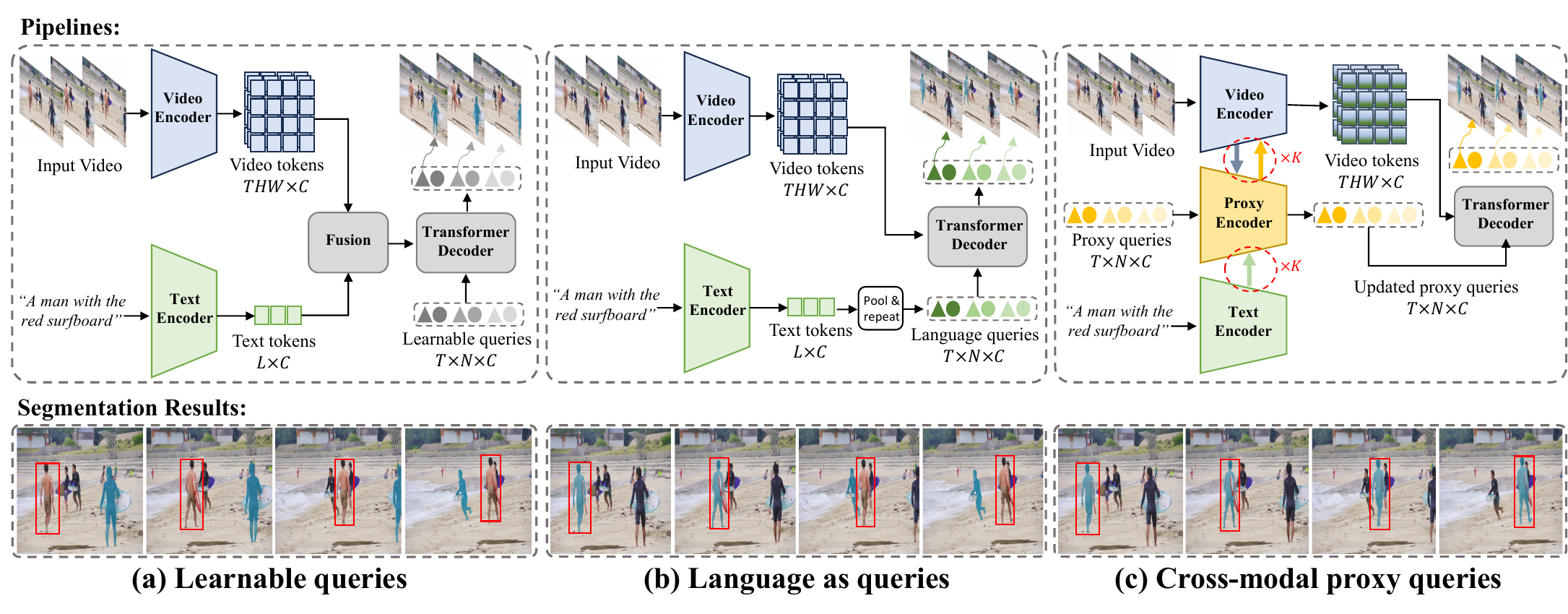}}
	%\vspace{-9pt}
	\caption{Comparison of current RVOS pipelines based on conditional queries and their segmentation results. (a) Learnable queries. (b) Language as queries. (c) Our cross-modality proxy queries. Bottom: The language expression for the target instance is \textit{"a person with the red surfboard"} and the target segmentation objects are indicated by the red boxes. Learnable queries method, MTTR \cite{DBLP:conf/cvpr/BotachZB22}, may segment incorrect objects due to the absence of linguistic constraints during the decoding stage. Language as queries based method, ReferFormer \cite{DBLP:conf/cvpr/WuJSYL22}, faces challenges in accurately tracking the target through significant frame-to-frame variations, due to a lack of modeling for inter-frame dependence and variability. Our cross-modality proxy queries based method, ProxyFormer, can produce more reasonable predictions.} 
	\label{fig:motiva}
%\vspace{-8pt}
\end{figure*}

Significant research efforts have been dedicated to exploring multi-modality segmentation techniques within the realms of audio-visual video object segmentation (AVOS) \cite{DBLP:conf/mm/LiYCY023, DBLP:conf/eccv/ZhouWZSZBGKWZ22} and referring video object segmentation (RVOS) \cite{DBLP:conf/cvpr/WuJSYL22, DBLP:conf/cvpr/BotachZB22}. During the early phase of RVOS, many methods \cite{DBLP:journals/corr/abs-2010-00263, DBLP:conf/eccv/SeoLH20} primarily focused on recognizing the object by implementing image-level strategies \cite{DBLP:conf/iccv/DingLWJ21, DBLP:conf/cvpr/LuoZSCWDJ20} on individual video frames, subsequently connecting them using heuristic rules. A clear limitation of these methods is their inability to utilize vital temporal information across frames, resulting in inconsistent object predictions due to appearance changes. 

%Leveraging the query-based mechanism of Transformers in detection \cite{DBLP:conf/eccv/CarionMSUKZ20} and segmentation \cite{DBLP:conf/cvpr/WangXWSCSX21}, recent RVOS approaches \cite{DBLP:conf/cvpr/WuD0S22, DBLP:conf/cvpr/BotachZB22, DBLP:conf/cvpr/WuJSYL22} utilize conditional queries as object candidates and feed them into the mask decoder.
As query-based mechanism in transformers has achieved marvelous success across detection \cite{DBLP:conf/eccv/CarionMSUKZ20} and segmentation \cite{DBLP:conf/cvpr/WangXWSCSX21} tasks, recent RVOS methods employ conditional queries as candidate objects, feeding them into a mask decoder to predict and track referred objects.
Conditional queries generally fall into two types: \textbf{(1) Learnable queries:} The typical learnable queries based methods (\textit{e.g}., MTTR \cite{DBLP:conf/cvpr/BotachZB22} and MLRL \cite{DBLP:conf/cvpr/WuD0S22}) query the video sequences for object-related information and retain it within the learnable queries in the decoder stage (see in Pipelines of Figure \ref{fig:motiva} (a)). 
They perform visual-language interaction for each frame separately and fail to incorporate text constraints into the learnable queries. This practice hinders the effective exploitation of inter-frame relationships and the comprehension of textual expressions detailing temporal variations of objects, potentially leading to segmentation errors.
For instance, in the bottom of Figure \ref{fig:motiva} (a), the referred expression is \textit{"a person with the red surfboard"} and the target segmentation objects are  marked by the red boxes. However, due to the absence of language constraints during the decoding stage, previous methods might segment the wrong objects.
\textbf{(2) Language as queries:} The recent language as queries based methods (\textit{e.g}., ReferFormer \cite{DBLP:conf/cvpr/WuJSYL22} and HTML \cite{DBLP:conf/iccv/00020L0C023}) utilize conditional queries (\textit{i.e.}, a set of object queries conditioned on the language expression) and obligate them to exclusively focus on the referred object (see in Pipelines of Figure \ref{fig:motiva} (b)). These approaches incorporate text restrictions and offer object-level guidance information during the decoding phase, significantly enhancing object segmentation accuracy. 
However, they handle the temporal feature of a video during the encoder stage and build visual-language interaction during the decoder stage sequentially, introducing the text constraints belatedly or insufficiently and leading to the segmentation results not complying with the text directives. 
Furthermore, these conditional queries lack modeling for inter-frame dependence and variability, with the same object across different frames being identified using the same conditional query, making it challenging to accurately track the target during significant frame-to-frame variations.
For example, in  the bottom of Figure \ref{fig:motiva} (b), due to the lack of insights on the video's temporal content for language conditioned queries, when multiple objects appear in the video, the target segmentation objects may drift at a certain moment, as shown in the third frame, leading to incorrect segmentation. 
%In summary, recent conditional queries based RVOS methods still encounter two primary limitations in the realms of 

Therefore, we propose a novel query-based approach, namely ProxyFormer, to address the challenge of visual-language alignment with cross-modality proxy queries. 
%and the temporal consistency of segmenting objects. 
Unlike previous works that introduce conditional queries at the decoder stage, we deploy our proxy queries during the video temporal modeling phase. The decision to place proxy queries early has two main motivations: (1) to serve as a proxy in transferring textual semantics to the video, ensuring that video features pay closer attention to the referred regions during the spatio-temporal modeling process, and (2) to progressively aggregate video and text semantics, including the temporal semantics, into proxy queries, generating stronger object-related queries for  predicting object masks. 
This dynamic evolution of the queries across video frames allows the model to better capture the object’s temporal variations and inter-frame dependencies.
Specifically, we propose a Cross-Modality Interaction Encoding (CMIE) based method, which contains two main components: Proxy Conditioned Video Encoding  and Visual-Language Conditioned Proxy Encoding. 
In video encoding, by progressively propagating proxy queries across multiple video encoder stages, it ensures that the video features are as focused as possible on the object of interest. 
In proxy encoding, we integrate the video and text semantics into proxy queries that efficiently aggregate intra-frame and inter-frame information of a video and capture the visual-language combined dependence. 
In summary, ProxyFormer’s progressive propagation of proxy queries addresses
these issues, dynamic changes in object appearance, multiple similar objects and temporal context confusion, by maintaining a coherent representation of the target object across frames and by simultaneously updating the interaction between video and text semantics, ensuring better alignment and tracking accuracy.

To mitigate the high computational costs caused by the full spatio-temporal interaction between video and proxy queries, we further decouple cross-modality interactions into their respective temporal and spatial dimensions. 
This approach streamlines the processing by handling each dimension individually, thereby enhancing efficiency without compromising the effectiveness of the cross-modality interaction.
Furthermore, we design a Joint Semantic Consistency (JSC) strategy in training  to align semantic consensus between the proxy queries and joint video-text pair and reduce the impact of semantic mismatches before mask prediction.
This strategy promotes a unified understanding, promoting a more accurate and compact integration of video and textual information.

To evaluate the effectiveness of our  method, we conduct extensive experiments on four popular RVOS benchmarks, \textit{i.e}., \textit{Ref-Youtube-VOS, Ref-DAVIS17, A2D-Sentences} and \textit{JHMDB-Sentences}. On \textit{Ref-Youtube-VOS} and \textit{Ref-DAVIS17}, ProxyFormer outperforms the baseline method ReferFormer \cite{DBLP:conf/cvpr/WuJSYL22} by +3.6\% and +4.2\% $\mathcal{J}\&\mathcal{F}$ scores, respectively, under the baseline Video-Swin-T setting. On  \textit{A2D-Sentences} and \textit{JHMDB-Sentences} ProxyFormer surpasses ReferFormer \cite{DBLP:conf/cvpr/WuJSYL22} by +5.0\% and +1.9\% mAP scores, respectively.

The key contributions of this work are summarized as follows:

$\bullet$
We propose a simple and new framework for referring video object segmentation, termed ProxyFormer, which  utilizes cross-modality proxy queries as intermediary mediums to (1) propagate and align visual-language semantics during encoding in a decoupled and memory-efficient manner, and (2) generate strongly object-related queries for predicting object masks during decoding.

$\bullet$
We further design a joint semantic consistency training strategy to align semantic consensus between the proxy queries and joint video-text pair and reduce the impact of semantic mismatches before mask prediction.

$\bullet$
%We conduct extensive experiments on four popular benchmarks, and achieve state-of-the-art segmentation performance.
We extensively evaluate the effectiveness of ProxyFormer on four popular RVOS benchmarks. Experimental results show that ProxyFormer achieves the state-of-the-art performance.

\section{Related Work}

\subsection{Vision-only Video Segmentation}
Vision-only video segmentation tasks generally encompass video object segmentation (VOS) \cite{DBLP:conf/cvpr/MiaoWWLL021, DBLP:conf/nips/YangY22a, DBLP:journals/tip/WangCGSS24, DBLP:journals/tmm/HongZFZ23}, video instance segmentation (VIS) \cite{DBLP:journals/tmm/LiWLL23, DBLP:journals/tmm/DaiCWPGS24, DBLP:conf/cvpr/WangXWSCSX21}, and video panoptic segmentation (VPS) \cite{DBLP:conf/cvpr/KimWLK20, DBLP:conf/cvpr/MiaoWWLZWY22}, all of which seek to accurately partition regions and annotate them with predefined categories or an open-world vocabulary \cite{DBLP:journals/corr/abs-2305-16835, DBLP:journals/corr/abs-2312-06703}. VOS aims to separate an object from the background within the video, based on the initial mask provided. VIS involves identifying and segmenting various instances within a specified category set while maintaining the unique identity of each instance throughout the tracking process. VPS targets the concurrent prediction of object classes, bounding boxes, masks, instance ID associations, and semantic segmentation, attributing unique identifiers to every pixel within a video.  
Recent advancements in Transformers \cite{DBLP:conf/nips/VaswaniSPUJGKP17} and DETR \cite{DBLP:conf/eccv/CarionMSUKZ20} have inspired the community towards end-to-end approaches, exemplified by works like VisTR \cite{DBLP:conf/cvpr/WangXWSCSX21} and SeqFormer \cite{DBLP:journals/corr/abs-2112-08275}.
Some approaches \cite{DBLP:conf/cvpr/Li0X023, DBLP:conf/cvpr/LiZQWLSJ21, DBLP:conf/eccv/WuJBZB22, DBLP:conf/eccv/WuLJBYB22} leverage the consistency of relative object relationships over short time spans to link entities across frames. 
The most recent advancements \cite{DBLP:conf/cvpr/HeoHHKOLK23, DBLP:conf/nips/HuangYA22, DBLP:conf/cvpr/LiLL021, DBLP:conf/cvpr/Li0XL0NS23} employ trainable trackers across multiple video clips, enhancing the learning of entity motion over short to medium duration.

\subsection{Referring Video Object Segmentation}
RVOS aims to segment and track target objects with the given natural language expressions in a video. In its early stages, most methods focused on identifying the object by applying image-level techniques  \cite{DBLP:journals/tmm/HuaLTZZ23, DBLP:journals/tmm/LiuJD23, DBLP:journals/tmm/DingZWYHCJ24, DBLP:conf/cvpr/LuoZSCWDJ20}  to individual video frames and then linking them through heuristic rules. 
A clear limitation of these methods is that they fail to leverage the crucial temporal information available across frames, leading to inconsistent object predictions owing to the variations in the scene or appearance. To address this issue, \cite{DBLP:journals/corr/abs-2106-01061}  introduced a top-down approach that involves creating a comprehensive set of object tracklets, followed by selecting the target through matching language features with all candidate tracklets. \cite{DBLP:conf/cvpr/WuD0S22} explored the inherent structure of video content to generate a set of discriminative visual embeddings, facilitating more effective alignment between vision and language semantics. MTTR \cite{DBLP:conf/cvpr/BotachZB22} formulated the RVOS task as a sequence prediction problem and generated predictions for all objects in the video before identifying the one specified by the text expression.
ReferFormer \cite{DBLP:conf/cvpr/WuJSYL22} also relied on the query-based mechanism and introduced a limited set of object queries that are conditioned on the text expression to focus solely on the referred object.

However, these methods still face two primary limitations in visual-language alignment and text-constrained decoding.
Unlike the previous methods \cite{DBLP:conf/cvpr/WuJSYL22, DBLP:conf/cvpr/BotachZB22} which handled temporal modeling and visual-language interaction as distinct elements, our proposed proxy queries, enriched with textual cues and object-level information, enable more accurate object segmentation compared to the previous  conditional queries based methods. 
%compared to the earlier model's conditional queries. 
By merging inter-frame dependency modeling with the synthesis of video and text semantics, our model achieves more precise results across four widely-used datasets.
%In this paper, our work continues to utilize the query-based paradigm but incorporates the alignment of visual-linguistic semantics during encoding through cross-modality proxy queries and applies stringent constraints to these proxy queries for predicting object masks during decoding. 
%Different from previous \cite{DBLP:conf/cvpr/WuJSYL22, DBLP:conf/cvpr/BotachZB22} methods based on conditional queries, our method based on proxy queries xxx.
%Unlike previous methods \cite{DBLP:conf/cvpr/WuJSYL22, DBLP:conf/cvpr/BotachZB22} based on conditional queries, which embed textual semantics into queries, our method relies on proxy queries to aggregate the joint semantics of video and text into the queries. Our approach not only propagates cross-modal semantics through the queries but also enables the queries to learn inter-frame dependencies during the process.

\section{Proposed Method}

\begin{figure*}[!h]
	\centering
	\centerline{\includegraphics[width=1\linewidth]{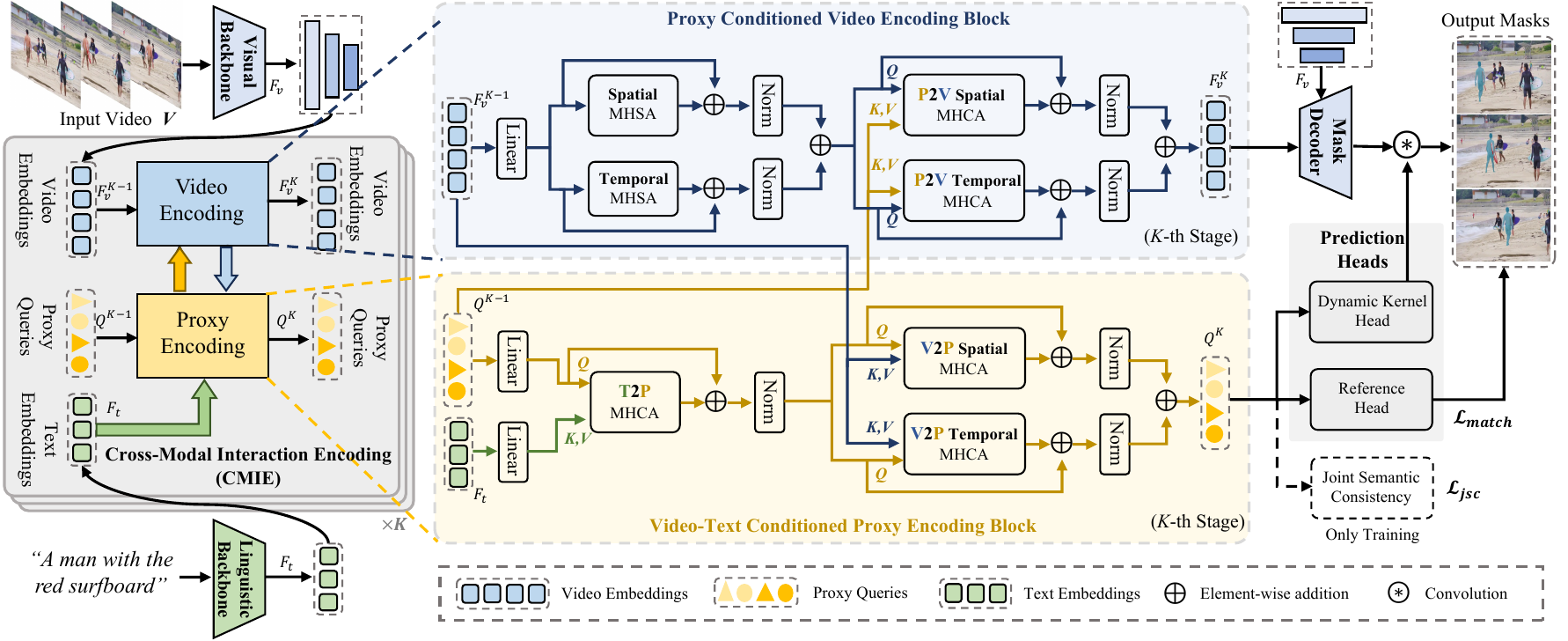}}
	%\vspace{-6pt}
	\caption{Overview of the proposed ProxyFormer framework. Given a frame sequence $\mathcal{V}= \{v_t\}_{t=1}^T$, and a textual expression $\mathcal{T}$, we first extract visual and linguistic features $F_v$ and $F_t$, then introduce a set of proxy queries $Q$ to bi-directionally interact with $F_v$ and $F_t$ via $K$ stacked Cross-Modality Interaction Encoding (CMIE) modules. Then these updated proxy queries $Q^K$ from the $K$-th CMIE module are used to predict object masks during decoding. Meanwhile, we design a Joint Semantic Consistency (JSC) training strategy to align semantic consensus between the proxy queries and the joint video-text pair.} 
	\label{fig:main}
%\vspace{-8pt}
\end{figure*}

%\subsection{Preliminary}
The input of RVOS consists of a frame sequence $\mathcal{V}= \{v_t\}_{t=1}^T$, $v_t \in \mathbb{R}^{H_0\times W_0 \times 3}$, and a text expression $\mathcal{T}= \{t_l\}_{l=1}^L$, with $t_l$ representing the $l$-th word in the text. RVOS aims to predict $T$-frame binary masks $\mathcal{M}= \{m_t\}_{t=1}^T$, $m_t \in \mathbb{R}^{H_0\times W_0}$, for the target object referred by $\mathcal{T}$. 
The overview of our proposed ProxyFormer is shown in Figure \ref{fig:main}, primarily consisting of four main components: (1) Visual and Language backbones, (2) Cross-Modality Interaction Encoding Modules (3) Mask Decoder and (4) Prediction Heads.
A set of proxy queries is first introduced into the Cross-Modality Interaction Encoding to incorporate video and text semantics into the proxy queries and enable a two-way flow of semantics between video and these proxy queries. Then, these proxy queries are fed into the mask decoder to identify the referred object. 

\noindent\textbf{Visual Backbone.} For a video sequence $\mathcal{V}$, visual features are extracted using popular backbones (\textit{e.g}., ResNet-50 and Video Swin Transformer). We denote the video features as $F_v \in \mathbb{R}^{T \times H\times W \times C_v}$, where $T$, $H$, $W$ and $C_v$ represent the number of frames, height, width, and channels, respectively.

\noindent\textbf{Language Backbone.} For a text expression  $\mathcal{T}$, the linguistic features are extracted using an off-the-shelf model, RoBERTa \cite{DBLP:journals/corr/abs-1907-11692}, which was pre-trained on multiple English-language corpora \cite{DBLP:journals/corr/abs-1806-02847, DBLP:conf/iccv/ZhuKZSUTF15}.
We denote the linguistic features as $F_t \in \mathbb{R}^{L \times C_t}$, where $L$ and $C_t$ represent the number of words and channels, respectively.

\noindent\textbf{Cross-Modality Interaction Encoding.} We introduce $N$ proxy queries to predict information related to objects for each frame, resulting in a total of $T\times N$ proxy queries for $T$ frames in a video. Each object query corresponds to a potential instance, with the same query across multiple frames being trained to represent the same instance throughout the video. 
%This mechanism facilitates the natural tracking of each instance throughout the video.  
To ensure that video features are maximally concentrated on the object of interest, we construct cross-modality semantics alignment via proxy conditioned video encoding and visual-language conditioned proxy encoding blocks.

\noindent\textbf{Mask Decoder.} In the mask decoder, we integrate the video features from the Cross-Modality Interaction Encoding module with multi-layer features from the backbone through an FPN-like architecture \cite{DBLP:conf/cvpr/WuJSYL22}, resulting in semantically-rich video feature maps used for producing segmentation masks.

\subsection{Cross-Modality Interaction Encoding}
Unlike previous works that introduce conditional queries at the decoder stage, we deploy a set of trainable queries, termed proxy queries, during the encoder phase. 
%The decision to place proxy queries early  is driven by  two main motivations: (1) to progressively aggregate video and text semantics, thereby generating strongly object-related queries for predicting object masks, and (2) to serve as proxy mediums for transferring textual semantics to the video, ensuring that video features pay closer attention to the referred regions during the spatio-temporal modeling process. 
We denote the initial proxy queries as $Q \in \mathbb{R}^{T \times N \times C_p}$, where $T$, $N$ and $C_p$ represent the number of frames, the number of object queries per frame, and channels, respectively.
For the video features $F_v \in \mathbb{R}^{T \times H\times W \times C_v}$, linguistic features $F_t \in \mathbb{R}^{L \times C_t}$ and proxy queries $Q \in \mathbb{R}^{T \times N \times C_p}$, we implement the $k$-th CMIE module as:
\begin{equation}\label{cmie}
F_v^k, Q^k = CMIE(F_v^{k-1}, F_t,  Q^{k-1}),
\end{equation}
where $CMIE(\cdot)$ denotes the Cross-Modality Interaction Encoding Module. 
In each CMIE module, we consistently inject linguistic features  $F_t$ from the language backbone to maintain the constraints of the referred expression. 
By stacking multiple CMIEs, we effectively extract high-quality target video features and capture the cross-modality aligned proxy queries. A CMIE module consists of two main components: Proxy Conditioned Video Encoding Block and Visual-Language Conditioned Proxy Encoding Block.
Due to the substantial memory requirements of using a transformer-based encoder, we adopt a spatio-temporal decoupling method to separately execute video encoding, proxy-to-video (P2V) interaction, and video-to-proxy (V2P) interaction.

\subsubsection{\textbf{Proxy Conditioned Video Encoding Block}}

The Proxy Conditioned Video Encoding Block is divided into two parts: Video Spatio-Temporal Encoding of the video itself and Proxy Conditioned Video Encoding.

\noindent
\textbf{Video Spatio-Temporal Encoding.} Instead of computing Multi-Headed Self-Attention (MHSA) across all tokens of video features $F_v^{k-1}$, we decouple the operation to compute self-attention in two parallel steps: spatially  (among all tokens extracted from the same temporal index by reshaping the tokens from $F_v^{k-1} \in \mathbb{R}^{THW \times C}$ to $F_{v,s}^{k} \in \mathbb{R}^{T \times HW \times C}$) and temporally (among all tokens extracted from the same spatial index by reshaping the tokens from $F_v^{k-1}$ to $F_{v,t}^{k} \in \mathbb{R}^{HW \times T \times C}$) in parallel.  The decoupled MHSA is defined as:
\begin{equation}
    \begin{aligned}\label{MHSA_st}
&\hat{F}_{v,s}^{k} = MHSA_s(F_{v,s}^{k}) + F_{v,s}^{k}, \\
&\hat{F}_{v,t}^{k} = MHSA_t(F_{v,t}^{k}) + F_{v,t}^{k},\\
&\hat{F}_{v}^{k} = \hat{F}_{v,s}^{k} + \hat{F}_{v,t}^{k}.  
    \end{aligned}
\end{equation}

\begin{figure}[!t]
	\centering
         %\vspace{-3pt}
	\centerline{\includegraphics[width=1\linewidth]{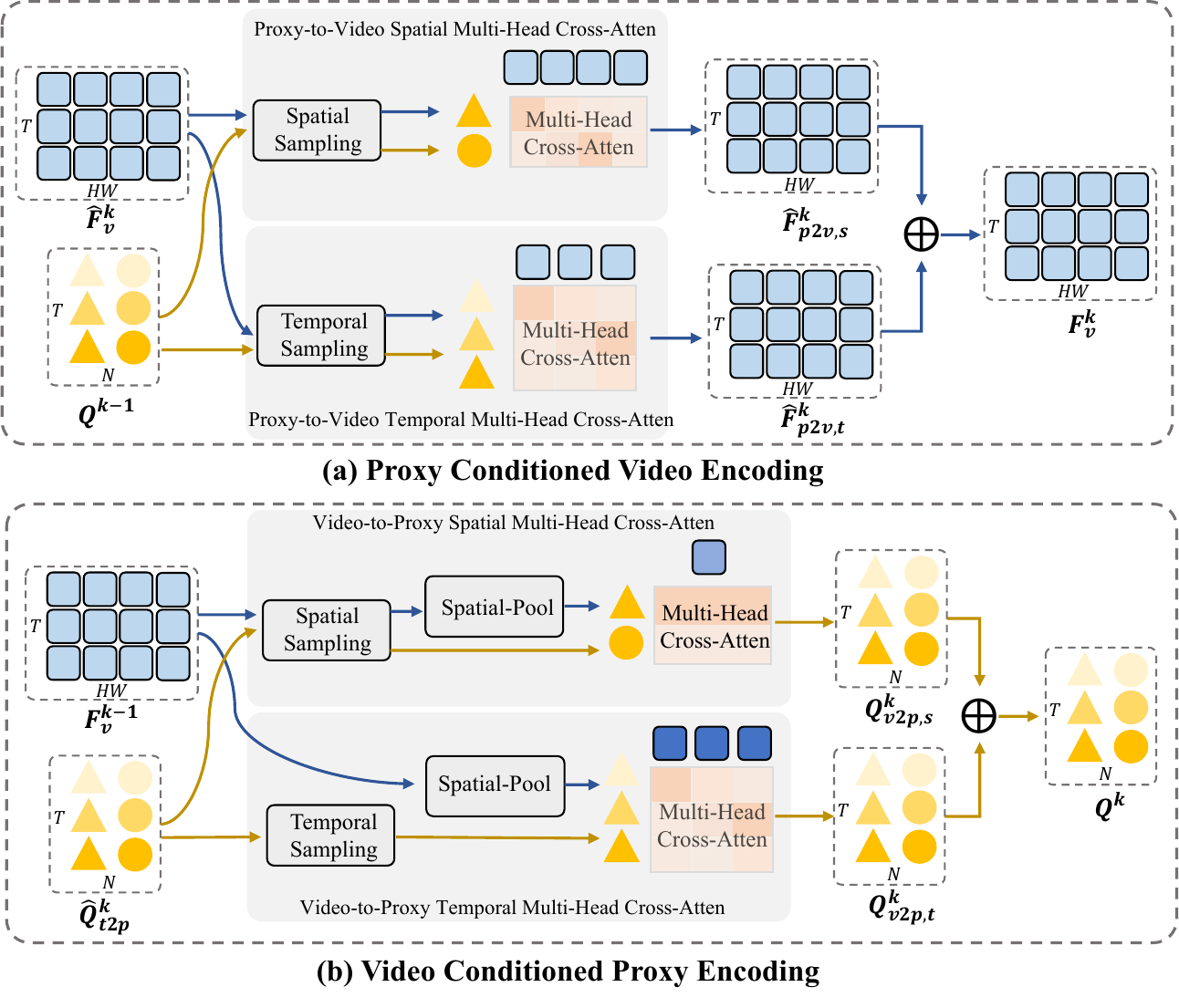}}
	%\vspace{-6pt}
	\caption{Illustration of the spatio-temporal divided cross-modality interaction of (a) proxy conditioned video encoding and (b) video conditioned proxy encoding.} 
	\label{fig:block}
%\vspace{-8pt}
\end{figure}

\noindent
\textbf{Proxy Conditioned Video Encoding.} As shown in Figure \ref{fig:block} (a), the video features $\hat{F}_{v}^{k}$ interact with proxy queries $Q^{k-1}$ through  Multi-Headed Cross-Attention (MHCA), where the \textit{query}  is $\hat{F}_{v}^{k}$, the \textit{key} and \textit{value} are  $Q^{k-1}$. For spatial proxy-to-video encoding, the video tokens are transformed from  $\hat{F}_{v}^{k} \in \mathbb{R}^{THW \times C}$ to ${F}_{p2v,s}^{k} \in \mathbb{R}^{T \times HW \times C}$, the proxy queries is $Q_{s}^{k} \in \mathbb{R}^{T \times N \times C}$.
For temporal proxy-to-video encoding, the video tokens are transformed from  $\hat{F}_{v}^{k}$ to ${F}_{p2v,t}^{k} \in \mathbb{R}^{HW \times T \times  C}$. We pool the proxy queries $Q^{k-1} \in \mathbb{R}^{T \times N \times C}$ for each frame and replicate it for $HW$ times to match the spatial resolution, obtaining $Q_{t}^{k} \in \mathbb{R}^{HW \times T \times C}$. The proxy conditioned video encoding is defined as follows:
\begin{equation}\label{MHCA_st_p2v}
    \begin{aligned}
& \hat{F}_{p2v,s}^{k} = MHCA_{s,p2v}({F}_{p2v,s}^{k}, Q_{s}^{k}) + {F}_{p2v,s}^{k}, \\
& \hat{F}_{p2v,t}^{k} = MHCA_{t,p2v}({F}_{p2v,t}^{k}, Q_{t}^{k}) + {F}_{p2v,t}^{k}, \\
& F_{v}^{k} = \hat{F}_{p2v,s}^{k} + \hat{F}_{p2v,t}^{k}.
    \end{aligned}
\end{equation}

\subsubsection{\textbf{Visual-Language Conditioned Proxy Encoding Block}}
The Visual-Language Conditioned Proxy Encoding Block is divided into two parts: the Linguistic Conditioned Proxy Encoding  and the Video Conditioned Proxy Encoding, which aggregate the language and video information into proxy queries continuously.

\noindent
\textbf{Linguistic Conditioned Proxy Encoding.} The proxy queries $Q^{k-1} \in \mathbb{R}^{T \times N \times C}$ interact with linguistic features $F_{t} \in \mathbb{R}^{L \times C}$ through MHCA, where the \textit{query}  is $Q^{k-1}$, the \textit{key} and \textit{value} are  $F_{t}$. 
It's noteworthy that within each CMIE module, we inject initial linguistic features $F_{t}$ from the linguistic backbone to maintain strong referring semantic constraints for proxy queries. The dimension of  proxy queries $Q^{k-1}$ is transformed to ${TN \times C}$. The linguistic conditioned proxy encoding is defined as:
\begin{equation}\label{MHCA_st_t2p}
    \begin{aligned}
\widetilde{Q}_{t2p}^{k} = MHCA_{t2p}(Q^{k-1}, F_{t}) + Q^{k-1}. \\
    \end{aligned}
\end{equation}

\noindent
\textbf{Video Conditioned Proxy Encoding.} As shown in Figure \ref{fig:block} (b), the proxy queries $\widetilde{Q}_{t2p}^{k} \in \mathbb{R}^{TN \times C}$ further interact with video features $F_{v}^{k-1} \in \mathbb{R}^{THW \times C}$ through MHCA, where the \textit{query}  is $\widetilde{Q}_{t2p}^{k}$, the \textit{key} and \textit{value} are  $F_{v}^{k-1}$.  Following the proxy conditioned video encoding, we perform the transformation operations and then obtain the video tokens $\widetilde{F}_{v,s}^{k} \in \mathbb{R}^{T \times HW \times C}$ and the proxy queries $\widetilde{Q}_{v2p,s}^{k} \in \mathbb{R}^{T \times N \times C}$ for spatial video-to-proxy encoding, and obtain the video tokens $\widetilde{F}_{v,t}^{k} \in \mathbb{R}^{N \times T \times C}$ by spatial pooling and  replicating $N$ times to match queries number for temporal video-to-proxy encoding. 
The proxy queries are transformed to $\widetilde{Q}_{v2p,t}^{k} \in \mathbb{R}^{N \times T \times  C}$ for temporal video-to-proxy encoding. The video conditioned proxy encoding is defined as:
\begin{equation}\label{MHCA_st_v2p}
    \begin{aligned}
& {Q}_{v2p,s}^{k} = MHCA_{s,v2p}(\widetilde{Q}_{v2p,s}^{k}, \widetilde{F}_{v,s}^{k}) + \widetilde{Q}_{v2p,s}^{k}, \\
& {Q}_{v2p,t}^{k} = MHCA_{t,v2p}(\widetilde{Q}_{v2p,t}^{k}, \widetilde{F}_{v,t}^{k}) + \widetilde{Q}_{v2p,t}^{k}, \\
& Q^{k} = {Q}_{v2p,s}^{k} + {Q}_{v2p,t}^{k}.
    \end{aligned}
\end{equation}

\begin{figure}[!t]
	\centering
         %\vspace{-3pt}
	\centerline{\includegraphics[width=1\linewidth]{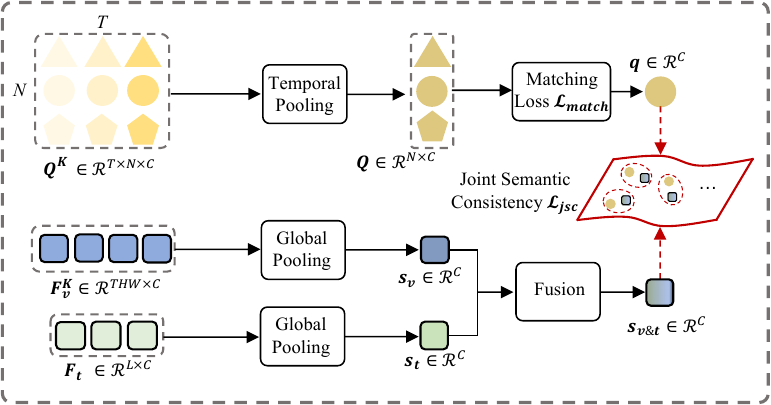}}
	%\vspace{-6pt}
	\caption{Illustration of the Joint Semantic Consistency (JSC). In JSC, we firstly pool  the proxy queries $Q^k$ in temporal dimension to generate video-level queries $Q_{v}$. Then, we find the best prediction as the positive video-level query $q$ according to the matching loss. Finally, we align semantic consensus between the video-level query $q$ and the joint video-text representation $s_{v\&t}$ through a joint semantic consistency loss.} 
	\label{fig:jsc}
%\vspace{-8pt}
\end{figure}

\subsection{Joint Semantic Consistency}
Several earlier studies \cite{DBLP:conf/bmvc/ChenTWLY19, DBLP:conf/cvpr/ChenGN18} aim to improve alignment by enforcing cyclical consistency between referring expressions and their respective reconstructions. However, they assume that the text and prediction semantics are identical or similar, which may not always hold true given the linguistic diversity and the inherent characteristics of the video itself.
In this section, we argue that the segmentation masks are jointly determined by both video and text, and relying solely on text semantics will narrow the prediction space, resulting in sub-optimal outcomes. 
Therefore, we design a Joint Semantic Consistency (JSC) training strategy to align semantic consensus between the proxy queries and the joint video-text pair.
In Figure \ref{fig:jsc}, we temporally pool  the proxy queries $Q^k$ from the last CMIE module to generate video-level queries $Q_{v} \in \mathbb{R}^{N \times C}$. Next, we identify the best prediction as the positive video-level query $q\in \mathbb{R}^{C}$ according to the matching loss:
\begin{equation}\label{jsc_match}
    \begin{aligned}
q & \leftarrow \mathop{argmin}\limits_{\hat{y}_i \in \hat{y}} \mathcal{L}_{match}(y, \hat{y}_i)
    \end{aligned}
\end{equation}
where $\mathcal{L}_{match}$ is defined as Eq. (\ref{matching}).

Then, we compute the video-text joint semantics $s_{v\&t}$ from the video features $F_v^k$ and text features $F_t$:
\begin{equation}\label{s_vt}
    \begin{aligned}
s_{v\&t} = \psi([GAP(F_v^k) + GAP(F_t)]), s_{v\&t} \in \mathbb{R}^{C},
    \end{aligned}
\end{equation}
where $GAP(\cdot)$ denotes global average pooling operation, $\psi$ is a linear transformation layer, $[\cdot]$ is a concatenation operation. 

Finally, we align semantic consensus between the video-level query $q$ and joint video-text representation $s_{v\&t}$ through a joint semantic consistency loss. This loss is defined using a symmetric contrastive loss, as shown below:
\begin{equation}\label{jsc}
    \begin{aligned}
\mathcal{L}_{jsc} = -\frac{1}{2\|\mathcal{B}\|}\sum\limits_{i=1}\limits^{\|\mathcal{B}\|}(\underbrace{log\frac{e^{x^{i}\cdot y^{i}}}{\sum_{j=1}^{\|\mathcal{B}\|}e^{x^{i}\cdot y^{j}}}}\limits_{q\rightarrow s_{v\&t}} + \underbrace{log\frac{e^{x^{i}\cdot y^{i}}}{\sum_{j=1}^{\|\mathcal{B}\|}e^{x^{j}\cdot y^{i}}}}\limits_{s_{v\&t}\rightarrow q}),
    \end{aligned}
\end{equation}
where $\mathcal{B}$ is the number of video of a batch. $x^{i}$ and  $y^{i}$ denote video-level query $q^i$ and joint video-text representation $s_{v\&t}^i$, respectively.

\subsection{Training Objectives}
\noindent\textbf{Dynamic Kernel Head.}
We first merge the final video feature $F_v^k$ and the multi-layered features from the visual backbone in a hierarchical manner using an FPN-like decoder, resulting in semantically-rich video feature maps $F_{seg}$. 
Then, we employ a two-layer perceptron, $\mathcal{G}_{kernel}$, to generate a  sequence of  dynamic kernels $\mathcal{K}$ from a series of $T\times N$ proxy queries. Ultimately, a sequence of segmentation masks $\mathcal{M}$ is produced by convolving each dynamic kernel with the respective frame features, which is then followed by bilinear upsampling to adjust the masks to the resolution of the ground truth,
\begin{equation}\label{mask}
    \begin{aligned}
\mathcal{M} = Upsampling(\mathcal{K}*F_{seg}),  \mathcal{M} \in \mathbb{R}^{T\times  N\times H_0\times W_0}.
    \end{aligned}
\end{equation}

\noindent\textbf{Reference Head.} 
The reference head comprises a classification head and a box head. The classification head directly predicts class probability $\hat{p} \in \mathbb{R}^{T\times  N \times N_{class}}$, where $N_{class}$ represents the number of classes. 
It's important to note that if $N_{class}=1$, the function of the classification head is to determine whether the object is referenced by the text expression. The box head is designed to transform the proxy queries into normalized bounding box information $\hat{b} \in \mathbb{R}^{T\times  N \times 4}$, including center coordinates, width, and height.

\noindent\textbf{Instance Matching.}
The prediction heads output the predicted trajectories of objects, denoted as $\hat{y}=\{\hat{y}_n\}$, with the prediction for the $n$-th object's trajectory being $\hat{y}_n=\{\hat{p}_n^t, \hat{b}_n^t, \hat{m}_n^t\}$. Given that there is only one referred object in the video, the ground-truth instance sequence is represented as $y=\{{p}^t, {b}^t, \hat{m}^t\}_t^T$.
We identify the best prediction as the positive sample $\hat{y}_{pos}$ by minimizing the matching cost:
\begin{equation}\label{matching}
    \begin{aligned}
& \hat{y}_{pos} = \mathop{argmin}\limits_{\hat{y}_i \in \hat{y}} \mathcal{L}_{match}(y, \hat{y}_n),\\
&\mathcal{L}_{match}(y, \hat{y}_n)= \lambda_{cls}\mathcal{L}_{cls} + \lambda_{box}\mathcal{L}_{box} + \lambda_{mask}\mathcal{L}_{mask},
    \end{aligned}
\end{equation}
where $\mathcal{L}_{cls}$ is focal loss that supervises the predicted object category, $\mathcal{L}_{box}$ is the L1 loss that supervises the regressed object box, $\mathcal{L}_{mask}$ is a combination of Dice loss and binary focal loss that supervises the predicted object mask, $\lambda$ is the scale factor used to balance each loss.

\noindent\textbf{Total Loss.} Our ProxyFormer is trained using the following total loss function that is a combination of the matching loss in Eq. (\ref{matching}) and the joint semantic consistency loss in Eq. (\ref{jsc}):
\begin{equation}\label{total_loss}
    \begin{aligned}
\mathcal{L}_{total} = \mathcal{L}_{match} + \lambda_{jsc}\mathcal{L}_{jsc}.
    \end{aligned}
\end{equation}

\section{Experimental Results and Discussions}

\subsection{Datasets and Evaluation Metrics}
\noindent\textbf{Dataset.} We train and validate our ProxyFormer on four RVOS benchmarks: Ref-Youtube-VOS \cite{DBLP:conf/eccv/SeoLH20}, Ref-DAVIS17 \cite{DBLP:conf/accv/KhorevaRS18}, A2D-Sentences  and JHMDB-Sentences \cite{DBLP:conf/mm/ChenLZYWH21}. Ref-Youtube-VOS \cite{DBLP:conf/eccv/SeoLH20} is a large-scale benchmark that encompasses 3,978 videos with around 15k language expressions. Ref-DAVIS17 \cite{DBLP:conf/accv/KhorevaRS18} is constructed on the foundation of DAVIS17 \cite{DBLP:journals/corr/Pont-TusetPCASG17} by offering language expression for specific objects in each video, containing a total of 90 videos. 
A2D-Sentences and JHMDB-Sentences \cite{DBLP:conf/mm/ChenLZYWH21} extend the original A2D \cite{DBLP:conf/cvpr/XuHXC15} and JHMDB \cite{DBLP:conf/iccv/JhuangGZSB13} datasets with additional text expressions. Specifically, A2D-Sentences comprises 3,754 videos, split into 3,017 for training and 737 for testing. JHMDB-Sentences features 928 videos, each with a text expression. 
%Typically, the entire dataset is used for evaluation after the RVOS model has been trained on A2D-Sentences.

\noindent\textbf{Evaluation Metrics.} We employ the standard metrics Jaccard index $\mathcal{J}$, F-score $\mathcal{F}$ and their average value $\mathcal{J}\&\mathcal{F}$ as evaluation metrics on Ref-Youtube-VOS and Ref-DAVIS17.
%, where $\mathcal{J}$ measures region similarity and $\mathcal{J}$ assesses contour accuracy, respectively.
%On A2D-Sentences and JHMDB-Sentences, the model is evaluated with Precision@K, Overall IoU, Mean IoU and mAP, where Precision@K measures the percentage of test samples whose IoU scores exceed the threshold K.
 For A2D-Sentences and JHMDB-Sentences, we use Precision@K, Overall IoU, Mean IoU, and mAP. Precision@K measures the percentage of test samples with IoU scores exceeding the threshold K.

\begin{table*}[!t]
	\begin{center}
	    %\vspace{-6pt}
		\caption{Comparison with the state-of-the-art methods on Ref-YouTube-VOS and Ref-DAVIS17.}
        %\vspace{-6pt}
        \label{table:comp_ref}
        \resizebox{0.78\textwidth}{!}{
		 \fontsize{8.0}{13}\selectfont
			\begin{tabular}{c|c|ccc|ccc}
				\toprule
				\multirow{2}{*}{\textbf{Method}}   & \multirow{2}{*}{\textbf{Backbone}}
				& \multicolumn{3}{c|}{\textbf{Ref-YouTube-VOS}}
				&\multicolumn{3}{c}{\textbf{Ref-DAVIS17}} \\
                \cline{3-8}&& \textbf{$\mathcal{J}\&\mathcal{F}$} (\%)&\textbf{$\mathcal{J}$ (\%)}&\textbf{$\mathcal{F}$} (\%)&\textbf{$\mathcal{J}\&\mathcal{F}$} (\%)&\textbf{$\mathcal{J}$} (\%)&\textbf{$\mathcal{F}$} (\%)\\
				\bottomrule
                %I3D \cite{DBLP:conf/cvpr/CarreiraZ17}  &K400&RGB&8&108G&64.4&75.6&28.2&66.7&33.2&48.3\\
                CMSA \cite{DBLP:conf/cvpr/YeR0W19} &ResNet-50&34.9 &33.3 &36.5&34.7 &32.2 &37.2\\
                %CMSA + RNN \cite{DBLP:conf/cvpr/YeR0W19} &ResNet-50& 36.4 &34.8 &38.1 &40.2 &36.9 &43.5\\
                URVOS \cite{DBLP:conf/eccv/SeoLH20} &ResNet-50&47.2 &45.3 &49.2 &51.5 &47.3 &56.0 \\
                LBDT-4 \cite{DBLP:conf/eccv/SeoLH20} &ResNet-50&47.2 &45.3 &49.2 &54.5 &- &-\\
                %MTTR \cite{DBLP:conf/cvpr/BotachZB22} &ResNet-50& & & & & & \\
                MLRL \cite{DBLP:conf/cvpr/WuD0S22}&ResNet-50&49.7 &48.4 &51.0 &58.0 &53.9 &62.0 \\
                HTML \cite{DBLP:conf/iccv/00020L0C023} &ResNet-50&57.8 &56.5 &59.0 &59.5 &56.6 &62.4 \\
                ReferFormer \cite{DBLP:conf/cvpr/WuJSYL22} &ResNet-50&55.6 &54.8 &56.5&58.5 &55.8 &61.3 \\
                \rowcolor{lightgray!60}ProxyFormer (Ours) &ResNet-50&\textbf{58.2} &\textbf{57.0} &\textbf{59.4} &\textbf{59.7} & \textbf{56.9}& \textbf{62.5}\\\hline 
                 MTTR \cite{DBLP:conf/cvpr/BotachZB22} &Video-Swin-T&55.3 &54.0 &56.6 &- &- &- \\
                %MLRL \cite{DBLP:conf/cvpr/WuD0S22}&Video-Swin-T& & & & & & \\
                HTML \cite{DBLP:conf/iccv/00020L0C023} &Video-Swin-T&61.2 &59.5 &63.0 &- &- &-  \\
                ReferFormer \cite{DBLP:conf/cvpr/WuJSYL22} &Video-Swin-T&59.4 &58.0 &60.9 &59.7 &56.6 &62.6 \\
                SgMg \cite{DBLP:conf/iccv/MiaoB0M23} &Video-Swin-T&62.0 &60.4 &63.5 &61.9 &59.0 &64.8 \\
                {BIFIT}~\cite{BIFIT}  & Video-Swin-T & 62.3 &60.6 &64.0 &60.5 &56.9 &64.1  \\ 
              FTEA~\cite{FTEA}  & Video-Swin-T & 56.5&  55.0 & 58.0& - & - & - \\
               TCE-RVOS~\cite{TCE-RVOS} & Video-Swin-T&61.3 &59.8 &62.7  & - & - & -\\ 
                SOC \cite{DBLP:conf/nips/LuoXLLWTLY23} &Video-Swin-T&62.4 &61.1 &63.7 &63.5 &60.2 &66.7 \\
                \rowcolor{lightgray!60}ProxyFormer (Ours) &Video-Swin-T&\textbf{63.0} &\textbf{61.9} &\textbf{64.1} &\textbf{63.9} &\textbf{60.5} &\textbf{66.8}\\\hline 
                %MTTR \cite{DBLP:conf/cvpr/BotachZB22} &Video-Swin-S& & & & & & \\
                %MLRL \cite{DBLP:conf/cvpr/WuD0S22}&Video-Swin-S& & & & & & \\
                %HTML \cite{DBLP:conf/iccv/00020L0C023} &Video-Swin-S&61.4 &59.9 &62.9 &- &- &- \\
                %ReferFormer \cite{DBLP:conf/cvpr/WuJSYL22} &Video-Swin-S&60.1 &58.6 &61.6 &- &- &- \\
                %\rowcolor{lightgray!60}ProxyFormer (ours) &Video-Swin-S&\textbf{62.2} &\textbf{60.9} &\textbf{63.4} &- &- &-\\\hline 
                %MTTR \cite{DBLP:conf/cvpr/BotachZB22} &Video-Swin-B& & & & & & \\
                %MLRL \cite{DBLP:conf/cvpr/WuD0S22}&Video-Swin-B&49.7 &48.4 &51.0 &58.0 &53.9 &62.0 \\
                HTML \cite{DBLP:conf/iccv/00020L0C023} &Video-Swin-B& 63.4 &61.5 &65.2 &62.1 &59.2 &65.1\\
                ReferFormer \cite{DBLP:conf/cvpr/WuJSYL22} &Video-Swin-B&62.9 &61.3 &64.6&61.1 &58.1 &64.1 \\
                OnlineRefer ~\cite{wu2023onlinerefer}  & Video-Swin-B& 62.9 & 61.0 & 64.7 & 62.4 & 59.1 & 65.6   \\
                {SLVP}~\cite{SLVP}  & Video-Swin-B  & 64.4 &62.5 &66.3 &61.3 &57.6 &64.9 \\ 
                SgMg \cite{DBLP:conf/iccv/MiaoB0M23} &Video-Swin-B&65.7 &63.9 &67.4 &63.3 &60.6 &66.0 \\
                SOC \cite{DBLP:conf/nips/LuoXLLWTLY23} &Video-Swin-B&66.0 &64.1 &67.9 &64.2 &61.0 &67.4 \\
                \rowcolor{lightgray!60}ProxyFormer (Ours) &Video-Swin-B&\textbf{66.3} &\textbf{65.0} &\textbf{68.3} &\textbf{64.6} &\textbf{61.6} &\textbf{67.9}\\ 
                %ATN&ImageNet&RGB&ResNet-50&16&66G& \textbf{86.8}&\textbf{91.2}&\textbf{46.8}&\textbf{84.6}&\textbf{71.6}&\textbf{82.1}\\
                \bottomrule
                 %MDCN \cite{DBLP:conf/mm/SunYYWLW22} &ImageNet&RGB&16 &70G&86.6&\textcolor{blue}{\textbf{89.9}}&\textcolor{blue}{\textbf{48.0}}&83.4
                %&\textcolor{blue}{\textbf{77.7}}&\textcolor{blue}{\textbf{85.3}}\\
               % MDCN \cite{DBLP:conf/mm/SunYYWLW22} &ImageNet&RGB&16 &70G&86.6&\textcolor{blue}{\textbf{89.9}}&\textcolor{blue}{\textbf{48.0}}&83.4
                %&\textcolor{blue}{\textbf{77.7}}&\textcolor{blue}{\textbf{85.3}}\\\hline \hline
                %\rowcolor{lightgray!40} ATNet (Ours) &ImageNet&RGB&16&72G & \textcolor{red}{\textbf{87.8}}&\textcolor{red}{\textbf{91.2}}&\textcolor{red}{\textbf{48.8}}&\textcolor{red}{\textbf{84.4}}&\textcolor{red}{\textbf{77.9}}&84.8\\
                %\bottomrule
		\end{tabular}}
	\end{center}
%\vspace{-8pt}
\end{table*}

\subsection{Implementation Details}
We build the proposed ProxyFormer upon the state-of-the-art pipeline, ReferFormer \cite{DBLP:conf/cvpr/WuJSYL22}. For our visual backbone, we use the Video Swin Transformer, and for the linguistic backbone, we utilize the RoBERTa-base model \cite{DBLP:journals/corr/abs-1907-11692}.
%We use video swin transformer as our visual backbone and utilize the RoBERTa-base model \cite{DBLP:journals/corr/abs-1907-11692} as linguistic backbone.
During the training phase, we typically input $\omega=8$ frames  surrounding the annotated frame into the model.
Within the transformer architecture, we use 4 layers (\textit{i.e}., $K=4$) for the encoder and 3 layers for the decoder, setting the hidden dimension to $C=256$. 
Both the encoder and decoder layers in the transformer feature 8 attention heads. Additionally, we set the number of object queries to $N=5$ for each frame.
The training process spans 30 epochs, starting with an initial learning rate of $10^{-4}$ for Ref-YouTube-VOS and $5\times10^{-5}$for A2D-Sentences. The learning rate decays by 10. For data augmentation, we employ RandomResize and Horizontal Flip techniques. In terms of resolution, all frames are resized to 360$\times$640 for Ref-YouTube-VOS and to 320$\times$576 for A2D-Sentences. 
For Refer-YouTube-VOS, we utilize windows of $\omega=12$ consecutive annotated frames during training and employ full-length videos for evaluation. In the case of A2D-Sentences, we feed the model windows of $\omega=8$ frames, positioning the annotated target frame in the middle.
The coefficients for the loss functions are set as follows: $\lambda_{cls}=2$, $\lambda_{L1}=5$, $\lambda_{giou}=2$, $\lambda_{dice}=5$, $\lambda_{focal}=2$ and $\lambda_{jsc}=5$. 

\begin{table*}[]
	\begin{center}
	    %\vspace{-6pt}
		\caption{Comparison with the state-of-the-art methods on A2D-Sentences.}
         %\vspace{-4pt}
		 \fontsize{8.0}{13}\selectfont
		\label{table:comp_a2d}
		\scalebox{0.90}{
			\begin{tabular}{c|c|ccccc|cc|c}
				\toprule
				\multirow{2}{*}{\textbf{Method}}   & \multirow{2}{*}{\textbf{Backbone}}
				& \multicolumn{5}{c|}{\textbf{Precision}}
				&\multicolumn{2}{c|}{\textbf{IoU}}& \multirow{2}{*}{\textbf{mAP}} \\
                \cline{3-9}&& \textbf{P@0.5}&\textbf{P@0.6}&\textbf{P@0.7}&\textbf{P@0.8}&\textbf{P@0.9}&\textbf{Overall IoU} &\textbf{Mean IoU}\\
				\bottomrule
                %I3D \cite{DBLP:conf/cvpr/CarreiraZ17}  &K400&RGB&8&108G&64.4&75.6&28.2&66.7&33.2&48.3\\
               % Hu et al. & VGG-16 &34.8 &23.6 &13.3 &3.3 &0.1 &47.4 &35.0 &13.2\\
               %Gavrilyuk et al. & I3D &47.5 &34.7 &21.1 &8.0 &0.2 &53.6 &42.1 &19.8\\
               CMSA + CFSA \cite{DBLP:journals/pami/YeRLZW22} & ResNet-101 &48.7 &43.1 &35.8 &23.1 &5.2 &61.8 &43.2 &-\\
               %ACAN & I3D &55.7 &45.9 &31.9 &16.0 &2.0 &60.1 &49.0 &27.4\\
               %CMPC-V & I3D &65.5 &59.2 &50.6 &34.2 &9.8 &65.3 &57.3 &40.4\\
               ClawCraneNet \cite{DBLP:journals/corr/abs-2103-10702} & ResNet-50/101 &70.4 &67.7 &61.7 &48.9 &17.1 &63.1 &59.9 &-\\
               MTTR ($\omega = 8$) \cite{DBLP:conf/cvpr/BotachZB22}& Video-Swin-T &72.1 &68.4 &60.7 &45.6 &16.4 &70.2& 61.8 &44.7\\
               MTTR ($\omega = 10$)\cite{DBLP:conf/cvpr/BotachZB22} & Video-Swin-T &75.4 &71.2 &63.8 &48.5 &16.9 &72.0 &64.0 &46.1\\
               MTML \cite{DBLP:conf/iccv/00020L0C023} &Video-Swin-T &82.2 &79.2 &72.3 &55.3& 20.1 &77.6 &69.2 &53.4\\
               MTML \cite{DBLP:conf/iccv/00020L0C023} &Video-Swin-B &84.0 &81.5 &75.8 &59.2 &22.8 &79.5 &71.2 &56.7\\
                ReferFormer \cite{DBLP:conf/cvpr/WuJSYL22} & Video-Swin-T &82.8 &79.2 &72.3 &55.3 &19.3 &77.6 &69.6 &52.8\\
               ReferFormer \cite{DBLP:conf/cvpr/WuJSYL22} & Video-Swin-B &83.1 &80.4 &74.1 &57.9 &21.2 &78.6 &70.3 &55.0\\
                \hline 
                \rowcolor{lightgray!60}ProxyFormer (ours)  &Video-Swin-T&\textbf{83.0} &\textbf{80.2} &\textbf{72.8} &\textbf{56.4} &\textbf{20.7} &\textbf{78.1}  &\textbf{70.6} &\textbf{54.3}\\
                \rowcolor{lightgray!60}ProxyFormer (ours)  &Video-Swin-B&\textbf{84.6} &\textbf{82.2} &\textbf{76.6} &\textbf{60.4} &\textbf{23.4} &\textbf{80.1} &\textbf{71.9} &\textbf{57.8}\\
                %ATN&ImageNet&RGB&ResNet-50&16&66G& \textbf{86.8}&\textbf{91.2}&\textbf{46.8}&\textbf{84.6}&\textbf{71.6}&\textbf{82.1}\\
                \bottomrule
                 %MDCN \cite{DBLP:conf/mm/SunYYWLW22} &ImageNet&RGB&16 &70G&86.6&\textcolor{blue}{\textbf{89.9}}&\textcolor{blue}{\textbf{48.0}}&83.4
                %&\textcolor{blue}{\textbf{77.7}}&\textcolor{blue}{\textbf{85.3}}\\
               % MDCN \cite{DBLP:conf/mm/SunYYWLW22} &ImageNet&RGB&16 &70G&86.6&\textcolor{blue}{\textbf{89.9}}&\textcolor{blue}{\textbf{48.0}}&83.4
                %&\textcolor{blue}{\textbf{77.7}}&\textcolor{blue}{\textbf{85.3}}\\\hline \hline
                %\rowcolor{lightgray!40} ATNet (Ours) &ImageNet&RGB&16&72G & \textcolor{red}{\textbf{87.8}}&\textcolor{red}{\textbf{91.2}}&\textcolor{red}{\textbf{48.8}}&\textcolor{red}{\textbf{84.4}}&\textcolor{red}{\textbf{77.9}}&84.8\\
                %\bottomrule
		\end{tabular}}
	\end{center}
%\vspace{-8pt}
\end{table*}

\subsection{Comparison with the State-of-the-Art}

\noindent\textbf{Ref-YouTube-VOS and Ref-DAVIS17.} We compare our ProxyFormer with the state-of-the-art (SOTA) methods on Ref-Youtube-VOS and Ref-DAVIS17 in Table \ref{table:comp_ref}. 
Specifically, on Ref-Youtube-VOS, ProxyFormer, utilizing a ResNet-50 backbone, achieves a remarkable $\mathcal{J}\&\mathcal{F}$  score of 58.2\%, which not only surpasses the recently baseline method ReferFormer by \textbf{2.6\%} with the same ResNet-50 backbone but also outperforms versions of ReferFormer that leverage larger backbones such as Video-Swin-Tiny and Video-Swin-Base by \textbf{3.6\%} and \textbf{3.4\%}, respectively. This trend of surpassing previous state-of-the-art performances extends to the Ref-DAVIS17 dataset, further underscoring the effectiveness and efficiency of ProxyFormer in handling complex RVOS tasks. This advancement highlights ProxyFormer's ability to better interpret and integrate visual and linguistic cues for precise object segmentation and tracking across video frames.

%On Ref-Youtube-VOS, our method achieves a score of xx\% in $\mathcal{J}\&\mathcal{F}$ with a ResNet-50 backbone. This performance exceeds that of the recent SOTA  method ReferFormer by 2.2\% under the  backbone of ResNet-50. Furthermore, it outperforms  ReferFormer that use larger backbones, Video-Swin-Tiny and Video-Swin-Base, on $\mathcal{J}\&\mathcal{F}$  by 1.2\% and 3.3\%, respectively. On Ref-DAVIS17, our approach continues to achieve SOTA performance, clearly demonstrating our method's superiority.

\begin{table}[!t]
	\begin{center}
	    %\vspace{-2pt}
		\caption{Comparison with the state-of-the-art methods on JHMDB-Sentences. O-IoU and M-IoU represent Overall IoU and Mean IoU.}
       %\vspace{-6pt}
		 \fontsize{8.0}{13}\selectfont
		\label{table:comp_jhmdb}
		\scalebox{0.87}{
			\begin{tabular}{c|c|cc|c}
				\toprule
				\multirow{2}{*}{\textbf{Method}}   & \multirow{2}{*}{\textbf{Backbone}}  &\multicolumn{2}{c|}{\textbf{IoU}} &\multirow{2}{*}{\textbf{mAP}} \\
                \cline{3-4}&& \textbf{O-IoU} &\textbf{M-IoU}&\\
				\bottomrule
                %I3D \cite{DBLP:conf/cvpr/CarreiraZ17}  &K400&RGB&8&108G&64.4&75.6&28.2&66.7&33.2&48.3\\
                MTTR ($\omega = 8$) \cite{DBLP:conf/cvpr/BotachZB22} &Video-Swin-T &67.4 &67.9&36.6\\
                MTTR ($\omega= 10$) \cite{DBLP:conf/cvpr/BotachZB22}& Video-Swin-T &70.1 &69.8&39.2\\
                %ReferFormer† (ω = 6) [29] Video-Swin-T 39.1
                MTML \cite{DBLP:conf/iccv/00020L0C023} &Video-Swin-T &-&-&42.7\\
                MTML \cite{DBLP:conf/iccv/00020L0C023} &Video-Swin-B &-&-&44.2\\
                 ReferFormer \cite{DBLP:conf/cvpr/WuJSYL22} & Video-Swin-T &71.9 &71.0&42.2\\
                ReferFormer \cite{DBLP:conf/cvpr/WuJSYL22} & Video-Swin-B &73.0 &71.8&43.7\\
                \hline 
                \rowcolor{lightgray!60}ProxyFormer (ours) &Video-Swin-T &\textbf{72.9}&\textbf{71.7}&\textbf{44.1} \\
                \rowcolor{lightgray!60}ProxyFormer (ours) &Video-Swin-B &\textbf{74.2}&\textbf{72.8}&\textbf{45.9} \\
                %ATN&ImageNet&RGB&ResNet-50&16&66G& \textbf{86.8}&\textbf{91.2}&\textbf{46.8}&\textbf{84.6}&\textbf{71.6}&\textbf{82.1}\\
                \bottomrule
                 %MDCN \cite{DBLP:conf/mm/SunYYWLW22} &ImageNet&RGB&16 &70G&86.6&\textcolor{blue}{\textbf{89.9}}&\textcolor{blue}{\textbf{48.0}}&83.4
                %&\textcolor{blue}{\textbf{77.7}}&\textcolor{blue}{\textbf{85.3}}\\
               % MDCN \cite{DBLP:conf/mm/SunYYWLW22} &ImageNet&RGB&16 &70G&86.6&\textcolor{blue}{\textbf{89.9}}&\textcolor{blue}{\textbf{48.0}}&83.4
                %&\textcolor{blue}{\textbf{77.7}}&\textcolor{blue}{\textbf{85.3}}\\\hline \hline
                %\rowcolor{lightgray!40} ATNet (Ours) &ImageNet&RGB&16&72G & \textcolor{red}{\textbf{87.8}}&\textcolor{red}{\textbf{91.2}}&\textcolor{red}{\textbf{48.8}}&\textcolor{red}{\textbf{84.4}}&\textcolor{red}{\textbf{77.9}}&84.8\\
                %\bottomrule
		\end{tabular}}
	\end{center}
%\vspace{-4pt}
\end{table}

\noindent\textbf{A2D-Sentences and JHMDB-Sentences.} 
We extended our evaluation to include the A2D-Sentences and JHMDB-Sentences datasets, benchmarking our approach against the latest SOTA achievements, as detailed in Tables \ref{table:comp_a2d} and Table \ref{table:comp_jhmdb}. 
ProxyFormer not only matches but also establishes new SOTA benchmarks on both datasets, underscoring its robustness.
Specifically, on the A2D-Sentences dataset, ProxyFormer achieves superior performance, outperforming the previous SOTA benchmarks by \textbf{1.5} mAP points with the Video-Swin-T backbone and by \textbf{2.8} mAP points with the Video-Swin-B backbone. Additionally, it shows significant improvements in recall at Precision@0.9 (\textbf{1.4} and \textbf{2.2} points for two backbones). 
On the JHMDB-Sentences dataset, ProxyFormer sets new records, reaching mAP scores of 44.1 and 45.9 for the Video-Swin-T and Video-Swin-B backbones, respectively. 
ProxyFormer not only sets a new benchmark in terms of mAP scores but also shows superior performance in IoU metrics, with O-IoU scores of 72.9 and 74.2, and M-IoU scores of 71.7 and 72.8 for the Video-Swin-T and Video-Swin-B backbones, respectively.
These results distinctly highlight the superior performance of our method on different benchmarks.

%%%%%%%%
%Contrasting with the previous model that treated temporal modeling of video and visual-linguistic interaction as separate components, our approach to combinatorial dependence more effectively counters the referred object. Our proxy queries, enriched with textual cues and object-level information, offer more precise object segmentation than the prior model's conditional queries. By synthesizing inter-frame dependency modeling with the integration of video and text semantics, our model achieves more accurate outcomes in four popular datasets.

\begin{table}[!t]
	\begin{center}
	    %\vspace{-3pt}
		\caption{Ablation analysis  of the key components in ProxyFormer on Ref-YouTube-VOS.}
        %\vspace{-6pt}
		 \fontsize{8.0}{13}\selectfont
		\label{table:ab_core}
		\scalebox{0.88}{
			\begin{tabular}{cc|ccc}
				\toprule
				&\textbf{Method} &\textbf{$\mathcal{J} \& \mathcal{F}$} (\%)&\textbf{$\mathcal{J}$} (\%)&\textbf{$\mathcal{F}$} (\%)\\
				\bottomrule
               (1)&Baseline  &59.4 &58.0 &60.9 \\
               (2)&Baseline + CMIE  &62.0 &60.6 &63.2\\
               (3)&Baseline + CMIE$^*$ &62.3 &61.0 &63.6\\
               (4)&Baseline + CMIE w/o P2V  &60.5 &58.6 &62.7\\
               \rowcolor{lightgray!60} (5)&Baseline + CMIE + JSC  &\textbf{63.0 \textcolor{red}{(+3.6)}} &\textbf{61.9 \textcolor{red}{(+3.9)}} &\textbf{64.1 \textcolor{red}{(+3.1)}}\\
                \bottomrule
		\end{tabular}}
	\end{center}
%\vspace{-4pt}
\end{table}

\subsection{Ablation Studies}
\label{ablation}
In this section, we perform ablation experiments to demonstrate the impact of the main components. All ablation experiments are conducted on Ref-YouTube-VOS benchmark using Video-Swin-Tiny as the visual backbone.

\noindent\textbf{The Impact of Proxy Queries.} 
%The introduction of proxy queries constitutes the core design of our proposed ProxyFormer, executed by the Cross-Modality Interaction Encoding (CMIE) module.
The key novelty of our proposed ProxyFormer lies in the introduction of proxy queries, effectively executed by the Cross-Modality Interaction Encoding (CMIE) module. 
From the 3-$rd$ to the 4-$th$ row of the Table \ref{table:ab_core}, we implement two variants of CMIE.
(1) \textbf{CMIE$^*$} denotes that the one performs comprehensive spatio-temporal interaction between video and proxy queries without adopting spatio-temporal decoupling. 
Compared to our memory-efficient spatio-temporal decoupling manner, the performance advantage of CMIE$^*$ is not significant, and it results in a noticeable increase in computational overhead (as further detailed in Table \ref{table:ab_cost}).
(2) \textbf{CMIE w/o P2V} denotes that, within the CMIE module, we merely aggregate video and textual semantics into the proxy queries. This variant does not facilitate interactions from the proxy queries back to the video (P2V), which results in a $\mathcal{J} \& \mathcal{F}$ performance decrease of \textbf{1.5\%} compared to the configuration presented in the 2-$nd$ row (CMIE).
We attribute the significance of P2V interaction to the predominant use of textual semantics in the video temporal encoding process, where the textual semantics contained in the proxy queries ensure that the video features are focused as much as possible on the object of interest. 

Even when a video involves just a single referred object, targeting multiple potential regions proves beneficial given the diversity of video content. In Table \ref{table:ab_number_N}, we study the impact of the number of proxy queries for each frame. It can be observed that increasing the number of object queries enhances the model's ability to more efficiently distinguish relevant objects from a broader array of potential regions.
Performance peaks at $N = 5$ and starts to  slightly decline as the number of proxy queries increases. This observation suggests that an excess of queries may cause confusion in the object grouping process across frames.
In Table \ref{table:ab_number_K}, we study the impact of the number $K$ of CMIE. When $K$ increases to 3, the model's performance significantly improves. Further increases to 4 and 5 result in only minor performance gains. We select $K=4$ as the default setting.

\begin{figure*}[!h]
	\centering
	\centerline{\includegraphics[width=0.94\linewidth]{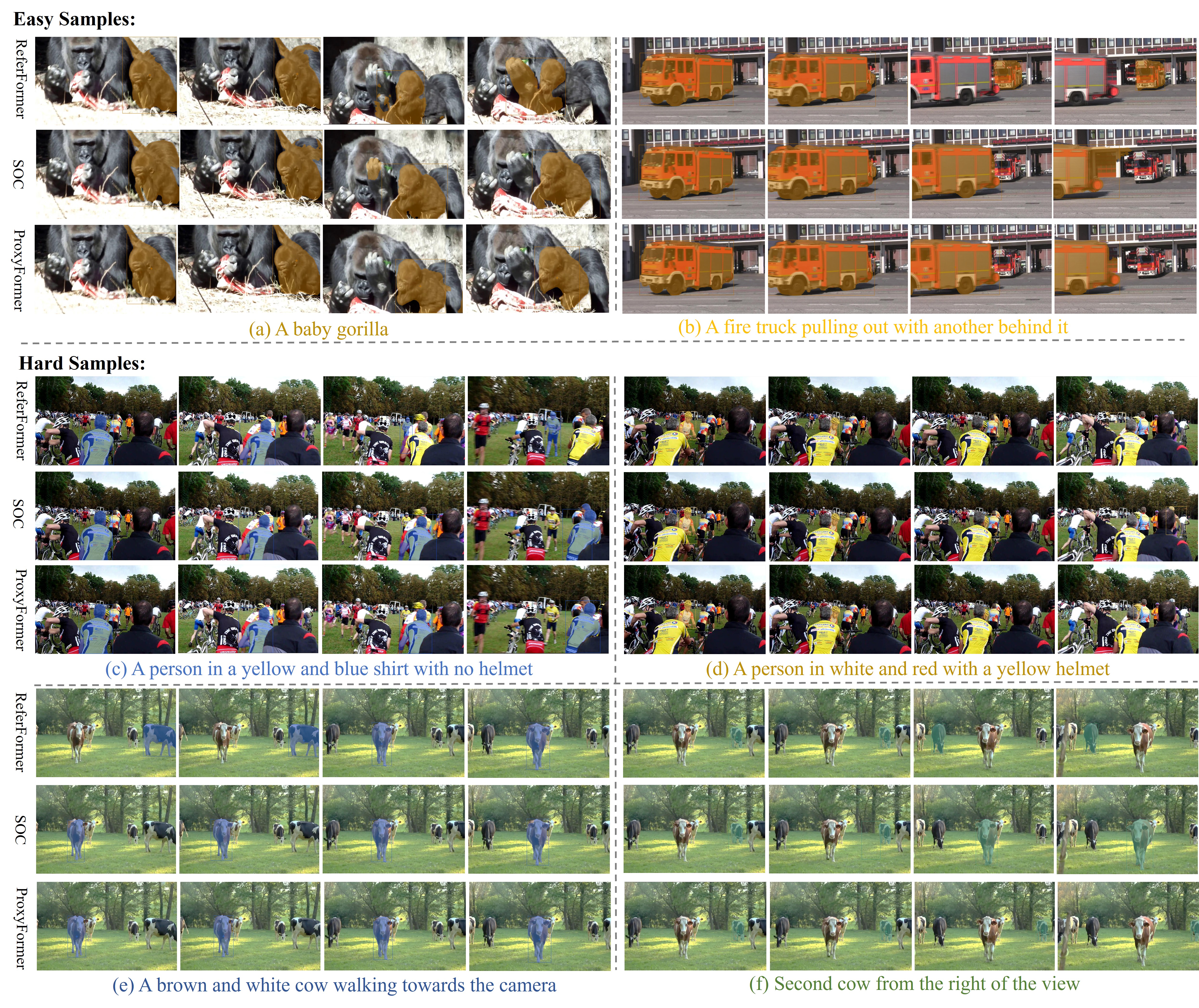}}
	%\vspace{-10pt}
	\caption{Qualitative comparison among our ProxyFormer, SOC \cite{DBLP:conf/nips/LuoXLLWTLY23} and ReferFormer \cite{DBLP:conf/cvpr/WuJSYL22} on Ref-YouTube-VOS. Our method effectively understands the spatial positions and appearances detailed in the queries, accurately identifying the referring objects. The colors of referring expressions correspond to the colors of segmentation masks.} 
	\label{fig:vis}
%\vspace{-8pt}
\end{figure*}

\begin{table}[!t]
	\begin{center}
	    %\vspace{-3pt}
		\caption{Ablation analysis  of the different semantic components (video semantic $s_v$ and text semantic $s_t$) in JSC.}
        %\vspace{-6pt}
		 \fontsize{8.0}{13}\selectfont
		\label{table:ab_jsc}
		\scalebox{0.9}{
		\begin{tabular}{cc|ccc}
				\toprule
             \textbf{video $s_v$} &\textbf{text $s_t$} &\textbf{$\mathcal{J} \& \mathcal{F}$} (\%)&\textbf{$\mathcal{J}$} (\%)&\textbf{$\mathcal{F}$} (\%)\\
				\bottomrule
              - & -&62.0 &60.6 &63.2\\
               \textbf{\checkmark} & -&60.3 &59.7 &62.2\\
               - &\textbf{\checkmark} &62.2 &61.2 &63.3\\
               \textbf{\checkmark}  & \textbf{\checkmark} &63.0 &61.9 &64.1\\
                \bottomrule
		\end{tabular}}
	\end{center}
%\vspace{-4pt}
\end{table}

\begin{table}[!t]
	\begin{center}
	    %\vspace{-3pt}
		\caption{Ablation analysis  of proxy query number $N$ on Ref-YouTube-VOS.}
        %\vspace{-6pt}
		 \fontsize{8.0}{13}\selectfont
		\label{table:ab_number_N}
		\scalebox{0.9}{
		\begin{tabular}{c|ccc}
				\toprule
				\textbf{Query Number} &\textbf{$\mathcal{J} \& \mathcal{F}$} (\%)&\textbf{$\mathcal{J}$} (\%)&\textbf{$\mathcal{F}$} (\%)\\
				\bottomrule
               \textbf{$N=1$}  &59.2 &58.9 &61.5\\
               \textbf{$N=3$}  &61.3 &59.6 &62.7\\
               \textbf{$N=5$}  &62.0 &60.6 &63.2\\
               \textbf{$N=7$}  & 62.1&60.9 &62.9\\
                \bottomrule
		\end{tabular}}
	\end{center}
%\vspace{-4pt}
\end{table}

\begin{table}[!t]
	\begin{center}
	    %\vspace{-3pt}
		\caption{Ablation analysis  of the number $K$ of CMIE on Ref-YouTube-VOS.}
        %\vspace{-6pt}
		 \fontsize{8.0}{13}\selectfont
		\label{table:ab_number_K}
		\scalebox{0.9}{
		\begin{tabular}{c|ccc}
				\toprule
				\textbf{CMIE Number} &\textbf{$\mathcal{J} \& \mathcal{F}$} (\%)&\textbf{$\mathcal{J}$} (\%)&\textbf{$\mathcal{F}$} (\%)\\
				\bottomrule
               \textbf{$K=2$}  &60.5&59.4 &62.0\\
               \textbf{$K=3$}   &61.6 &59.4 &62.9\\
               \textbf{$K=4$}   &62.0 &60.6 &63.2\\
               \textbf{$K=5$}   & 61.8&59.7 &63.6\\
                \bottomrule
		\end{tabular}}
	\end{center}
%\vspace{-8pt}
\end{table}

% \begin{table}[!t]
% 	\begin{center}
% 	    %\vspace{-4pt}
% 		\caption{Segmentation accuracy ($\mathcal{J} \& \mathcal{F}$ (\%)), computation cost (FLOPs) and inference time ${T}_{infer}$ (ms) for a sample.}
%         %\vspace{-6pt}
% 		 \fontsize{8.0}{13}\selectfont
% 		\label{table:ab_cost}
% 		\scalebox{1}{
% 			\begin{tabular}{c|ccc}
% 				\toprule
% 				\textbf{Method} &\textbf{$\mathcal{J} \& \mathcal{F}$} &\textbf{FLOPs} & ${T}_{infer}$ (ms) \\
% 				\bottomrule
%               \textbf{ReferFormer} \cite{DBLP:conf/cvpr/WuJSYL22}  &59.4 &1298.8G &104\\
%               %MTML \cite{DBLP:conf/iccv/00020L0C023} & & &\\
%               \textbf{ProxyFormer$^*$} &63.5 &1484.6G &120\\
%               \textbf{ProxyFormer}  &63.0 &1341.6G &109\\
%                 \bottomrule
% 		\end{tabular}}
% 	\end{center}
% \vspace{-4pt}
% \end{table}

\begin{table}[!t]
	\begin{center}
	    %\vspace{-4pt}
		\caption{Segmentation accuracy ($\mathcal{J} \& \mathcal{F}$ (\%)), computation cost (FLOPs) and inference time ${T}_{infer}$ (ms) for a sample.}
        %\vspace{-6pt}
		 \fontsize{8.0}{13}\selectfont
		\label{table:ab_cost}
		\scalebox{0.88}{
			\begin{tabular}{c|ccc}
				\toprule
				\textbf{Method} &\textbf{$\mathcal{J} \& \mathcal{F}$} &\textbf{FLOPs} & ${T}_{infer}$ (ms) \\
				\bottomrule
              \textbf{(a) Baseline}  &59.4 &1298.8G &104\\
              %MTML \cite{DBLP:conf/iccv/00020L0C023} & & &\\
              \textbf{(b) ProxyFormer (w/o proxy queries)} &61.0 &1756.0G &134\\
              \textbf{(c) ProxyFormer (full model)}  &63.0 &1341.6G &109\\
              \textbf{(d) ProxyFormer$^*$ (full model)} &63.5 &1484.6G &120\\
                \bottomrule
		\end{tabular}}
	\end{center}
%\vspace{-8pt}
\end{table}

\noindent\textbf{The Impact of Joint Semantic Consistency.} 
We conduct experiments to evaluate the effectiveness of the proposed Joint Semantic Consistency (JSC).  The last row of Table \ref{table:ab_core} presents the result of combining CMIE with JSC.  We observe a \textbf{1.0\%} $\mathcal{J} \& \mathcal{F}$  performance improvement  than the single CMIE. We further investigate the impact of different semantic components, \textit{i.e}., video semantics $s_v$, text semantics $s_t$ and their joint semantics, as shown in Table \ref{table:ab_jsc}.
Using video semantics $s_v$ alone results in a decrease in performance, while employing text semantics $s_t$ leads to a slight improvement. The combination of both video and text semantics yields even better results, with a significant increase of $\mathcal{J} \& \mathcal{F}$. This emphasizes the importance of achieving semantic consensus alignment between the video-level query and the joint video-text representation.

\noindent\textbf{Analysis of Computation Cost.}
We analyze the computation cost (FLOPs) and inference time ${T}_{infer}$ of ProxyFormer in Table \ref{table:ab_cost}. 
For a fair comparison, the inference times of our method and ReferFormer \cite{DBLP:conf/cvpr/WuJSYL22} (baseline model) were calculated under the same conditions, with the resolution of the input frames set at 360$\times$640 for a total of 19 frames, using an NVIDIA A40-48G GPU.
ProxyFormer (w/o proxy queries): This model includes the proxy encoder, but proxy queries are not used for cross-modality interaction. Instead of proxy queries mediating video-text alignment, video and text features directly interact through cross-attention at each stage.
ProxyFormer$^*$ denotes that the CMIE module performs comprehensive spatio-temporal interaction between video and proxy queries without adopting spatio-temporal decoupling.
When proxy queries are removed, but the proxy encoder remains, performance drops from 63.0\% to 61.0\% J\&F.
This version performs slightly better than the baseline, but at the cost of higher computation.
This confirms that proxy queries play a crucial role in bridging video and text semantics, and their absence leads to misalignment in feature fusion.
Compared to ProxyFormer*, our ProxyFormer significantly reduces computational costs and inference time, with only a modest decrease in segmentation accuracy (a reduction of 0.5\%).
Compared to ReferFormer, our ProxyFormer requires only an additional 3.3\% (+42.8 GFLOPs) computational cost and 5ms of inference time yet achieves a 3.6\% $\mathcal{J} \& \mathcal{F}$ improvement, which is acceptable considering the performance enhancement.

\subsection{Visualization Analysis}
In Figure 2, we provide a visualization comparison between the proposed ProxyFormer and the baselines, including ReferFormer \cite{DBLP:conf/cvpr/WuJSYL22}  and SOC \cite{DBLP:conf/nips/LuoXLLWTLY23}, illustrating the performance enhancements achieved by our method. 
For this visualization, we select Easy ((a) and (b)) and Hard ((c), (d), (e) and (f)) samples
respectively. The easy samples contain simple video scenes and referring texts, while the hard samples involve complex video scenes, such as multiple actors and severe occlusions. For the easy action samples, both ReferFormer, SOC and our ProxyFormer can locate the target objects, such as the baby gorilla and the fire truck. However, compared to ReferFormer and SOC, ProxyFormer provides more precise segmentation mask boundaries.
For certain hard samples, ReferFormer produces unreasonable predictions and loses the correct targets over time due to the lack of inter-frame dependencies for the segmented targets. Within the proposed proxy query-based framework, ProxyFormer is able to generate more reasonable predictions and effectively maintain the temporal consistency of the segmented targets.

\section{Conclusion}
In this paper, we present ProxyFormer, a novel RVOS method featuring proxy queries designed to merge video and text semantics and enable semantic exchange between them. By continuously refining and transmitting these queries through the video feature encoding stages, ProxyFormer ensures heightened focus on the target object and facilitates the establishment of inter-frame relationships.
To address the computational demands of extensive spatio-temporal interactions between video and proxy queries, we decouple cross-modality interactions into separate temporal and spatial components.
Additionally, we implement a joint semantic consistency training strategy to harmonize the semantic consensus between the proxy queries and the combined video-text entities.
Extensive experiments on four established RVOS benchmarks demonstrate ProxyFormer's superiority over the SOTA methods.

\bibliographystyle{IEEEtran}
\bibliography{mybib}

\begin{IEEEbiography}[{\includegraphics[width=1in,height=1.25in,clip,keepaspectratio]{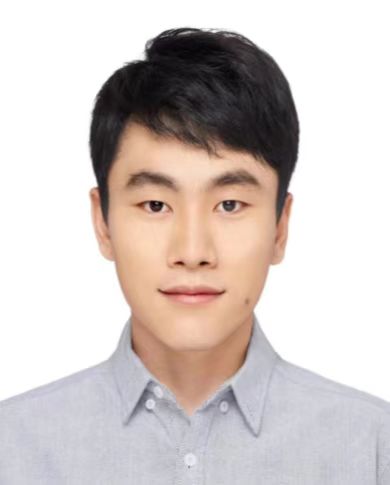}}]{Baoli Sun} is currently pursuing the Ph.D. degree with the School of Software Technology, Dalian University of Technology. He received the B.S degree in microelectronics science and engineering in 2018 from the Hefei University of Technology , Anhui , China.  He received his M.S. degree in software engineering in 2021 from the Dalian University of Technology, Dalian, China. His research interests include computer vision and deep learning.
\vspace{-10mm}
\end{IEEEbiography}

\begin{IEEEbiography}[{\includegraphics[width=1in,height=1.25in,clip,keepaspectratio]{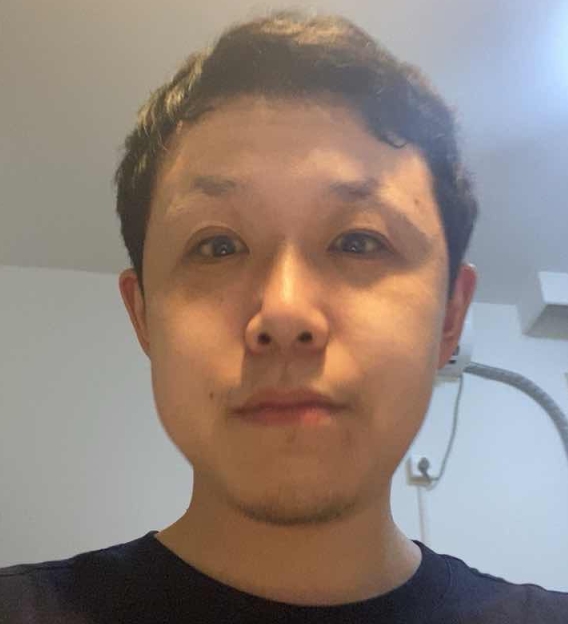}}]{Xinzhu Ma} received his B.Eng and M.P’s degree from Dalian University of Technology in 2017 and 2019, respectively. After that, He got the Ph.D degree from the University of Sydney in 2023. He is currently a postdoctoral researcher at the Chinese University of Hong Kong. His research interests include deep learning and computer vision.
\vspace{-10mm}
\end{IEEEbiography}

\begin{IEEEbiography}[{\includegraphics[width=1in,height=1.25in,clip,keepaspectratio]{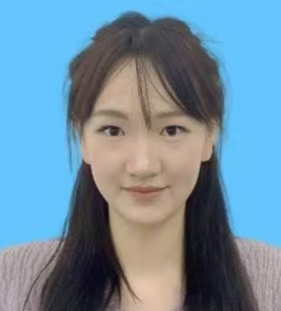}}]{Ning Wang} received a Ph.D. degree in Software of Engineering, at the Dalian University of Technology, China. She  received a B.Sc. degree from the software of engineering, at Dalian University of Technology. Her research interests include image colorization, video colorization, image processing, and multi-modal fusion.
\vspace{-10mm}
\end{IEEEbiography}

\begin{IEEEbiography}[{\includegraphics[width=1in,height=1.25in,clip,keepaspectratio]{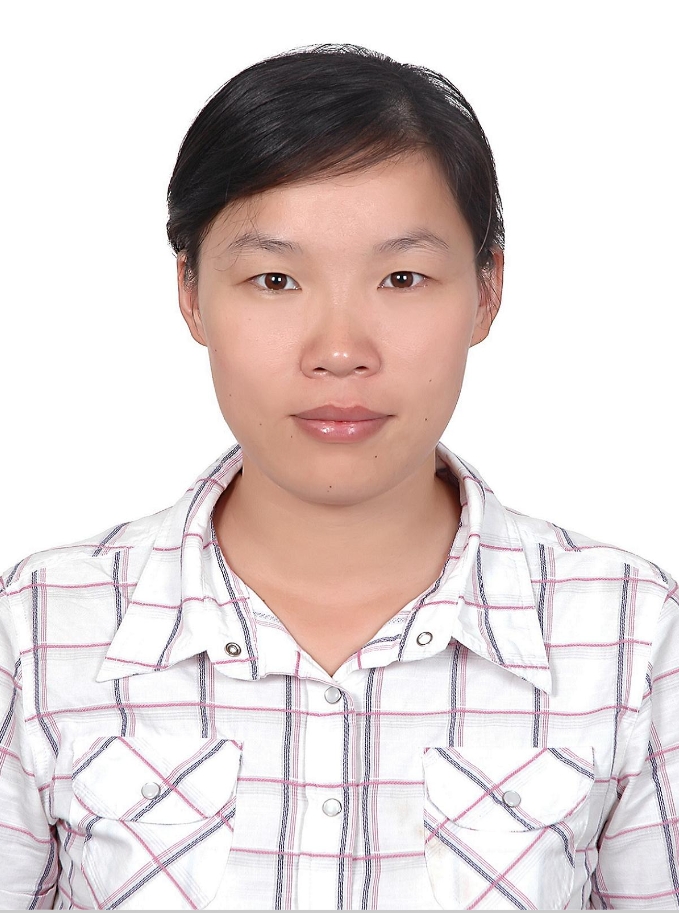}}]{Zhihui Wang}
received the B.S. degree in software engineering in 2004 from the North Eastern University, Shenyang, China. She received her M.S. degree in software engineering in 2007 and the Ph.D degree in software and theory of computer in 2010, both from the Dalian University of Technology, Dalian, China. Her current research interests include information hiding and image compression.
\vspace{-10mm}
\end{IEEEbiography}

\begin{IEEEbiography}[{\includegraphics[width=1in,height=1.25in,clip,keepaspectratio]{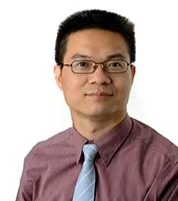}}]{Zhiyong Wang}
received his B.Eng. and M.Eng. degrees in electronic engineering from South China University of Technology, Guangzhou, China, and his Ph.D. degree from Hong Kong Polytechnic University, Hong Kong. He is a Senior Lecturer of the School of Information Technologies, the University of Sydney. His research interests include multimedia information processing, retrieval and management, human-centered multimedia computing.
\vspace{-10mm}
\end{IEEEbiography}

\end{document}